\documentclass[lettersize,journal]{IEEEtran}
\usepackage{amsmath,amsfonts}
\usepackage{algorithmic}
\usepackage{algorithm}
\usepackage{array}
\usepackage[caption=false,font=normalsize,labelfont=sf,textfont=sf]{subfig}
\usepackage{textcomp}
\usepackage{stfloats}
\usepackage{url}
\usepackage{verbatim}
\usepackage{graphicx}
\usepackage{cite}
\usepackage{makecell}
\usepackage{amsmath}
\usepackage{multirow}
\usepackage{subcaption}
\usepackage{caption}
\usepackage{newtxtext} % Times New Roman font
\usepackage{float}
\usepackage[T1]{fontenc} % Ensures that the output font encoding is set properly
\begin{document}

\title{Autonomous Robotic System with Optical Coherence Tomography Guidance for Vascular Anastomosis}

\author
{Jesse Haworth$^{1}$, Rishi Biswas$^{1}$, Justin Opfermann$^{1}$, Michael Kam$^{1}$, Yaning Wang$^{2}$, Desire Pantalone$^{3}$, \\Francis X. Creighton$^{4}$, Robin Yang$^{5}$, Jin U. Kang$^{2}$, and Axel Krieger$^{1}$,~\IEEEmembership{Member,~IEEE,}%
%\author{IEEE Publication Technology,~\IEEEmembership{Staff,~IEEE,}
        % <-this % stops a space

\thanks{*Research was supported by the National Institute of Biomedical Imaging and Bioengineering of the National Institutes of Health (NIH) under awards 1R01EY032127, 1R56EB033807 and the National Science Foundation (NSF) Foundational Research in Robotics under CAREER award 2144348. The content is solely the responsibility of the authors and does not necessarily represent the official views of the NIH or NSF.}% <-this % stops a space
\thanks{$^{1}$Jesse Haworth, Rishi Biswas, Justin Opfermann, Michael Kam, and Axel Krieger are with the Department of Mechanical Engineering, Johns Hopkins University, 3400 N. Charles St., Baltimore, MD, United States (email: jhawort2@jhu.edu; rbiswas4@jhu.edu; jopferm1@jhu.edu; mkam2@jhu.edu; axel@jhu.edu)}%
\thanks{$^{2}$Yaning Wang and Jin U. Kang are with the Department of Electrical Engineering, Johns Hopkins University, 3400 N. Charles St., Baltimore, MD, United States (email: ywang511@jhu.edu; jkang@jhu.edu)}%
\thanks{$^{3}$Desire Pantalone is with the Department of Clinical and Experimental Medicine, University of Florence, Florence, Italy and the Emergency Surgery Unit-Trauma Team, Trauma Center, Careggi University Hospital, Florence, Italy (email: desire.pantalone@unifi.it)}%
\thanks{$^{4}$Francis X. Creighton is with the Department of Otolaryngology - Head and Neck Surgery, Johns Hopkins School of Medicine, 720 Rutland Ave, Baltimore, MD, United States (email: francis.creighton@jhmi.edu)}%
\thanks{$^{5}$Robin Yang is with the Department Plastics and Reconstructive Surgery, Johns Hopkins School of Medicine, 720 Rutland Ave, Baltimore, MD, United States (email: ryang114@jhmi.edu)}%
}

\markboth{Journal of \LaTeX\ Class Files,~Vol.~14, No.~8, August~2021}%
{Shell \MakeLowercase{\textit{et al.}}: A Sample Article Using IEEEtran.cls for IEEE Journals}

\IEEEpubid{0000--0000/00\$00.00~\copyright~2021 IEEE}

\maketitle

\begin{abstract}

Vascular anastomosis, the surgical connection of blood vessels, is essential in procedures such as organ transplants and reconstructive surgeries. The precision required limits accessibility due to the extensive training needed, with manual suturing leading to variable outcomes and revision rates up to 7.9\%. Existing robotic systems, while promising, are either fully teleoperated or lack the capabilities necessary for autonomous vascular anastomosis. We present the Micro Smart Tissue Autonomous Robot (µSTAR), an autonomous robotic system designed to perform vascular anastomosis on small-diameter vessels. The µSTAR system integrates a novel suturing tool equipped with Optical Coherence Tomography (OCT) fiber-optic sensor and a microcamera, enabling real-time tissue detection and classification. Our system autonomously places sutures and manipulates tissue with minimal human intervention. In an ex vivo study, µSTAR achieved outcomes competitive with experienced surgeons in terms of leak pressure, lumen reduction, and suture placement variation, completing 90\% of sutures without human intervention. This represents the first instance of a robotic system autonomously performing vascular anastomosis on real tissue, offering significant potential for improving surgical precision and expanding access to high-quality care.
\end{abstract}

\vspace{3em}
\section{Introduction}
\IEEEPARstart{V}{ascular} anastomosis, the surgical connection of two blood vessels, is a critical procedure in numerous surgical disciplines, including organ transplants, reconstructive surgeries, and microvascular tissue transfers. In 2020, more than six million reconstructive surgeries were performed in the United States alone \cite{PlasticSurgeryStats2020}, many of which involved vascular anastomosis. The precision required for successful vascular anastomosis, particularly in vessels smaller than one millimeter in diameter, demands extensive training and experience. This requirement limits the number of surgeons capable of performing such procedures, thereby reducing access to these life-saving operations, especially in underserved regions. 

\begin{figure}[t]
    \centering
    \includegraphics[width=1.0\linewidth]{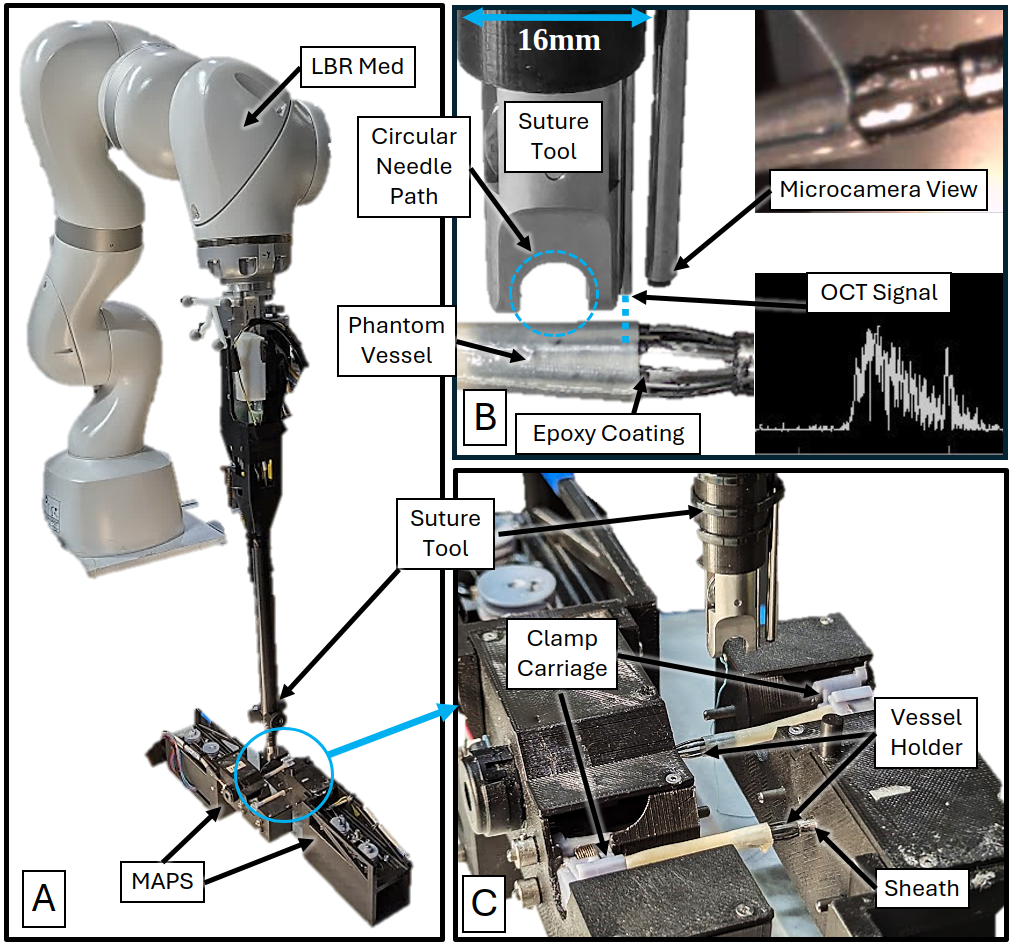}
    \caption{µSTAR System overview. A: LBR Med robotic manipulator and suture tool positioned over the microvascular anastomosis positioning system (MAPS). B: Suturing tool equipped with OCT fiber and microcamera. C: MAPS clamp carriage and nitinol vessel holder.}
    \label{fig:µSTARsystemoverview}
    \vspace{-2em}
\end{figure}

The current gold standard for vascular anastomosis is manual suturing, which is highly dependent on the surgeon's skill. Studies have reported revision rates as high as 7.9\% \cite{heidekrueger2022comparison}, with complications such as leakage, thrombosis, and stenosis often resulting from inconsistencies in suture placement \cite{goss2018fundamentals}\cite{saade2023tricks}. The anticipated shortage of 13,500 to 86,000 physicians by 2036 in the United States alone \cite{aamcglobaldata2024} underscores the urgent need for technological solutions that reduce dependence on individual surgical skill and enhance the consistency of clinical outcomes. 

Robotic systems are one such technology that have the  potential to enhance the precision and consistency of vascular anastomosis. The Da Vinci robot (Intuitive Surgical, Sunnyvale, CA) \cite{lai2019robot}, MUSA robot (Microsure, Eindhoven, The Netherlands) \cite{SymaniInVivoTrial2023}, and Symani Surgical System (Medical Microinstruments, Jacksonville, FL)  \cite{micosuremusatrial2022} offer the surgeon improved ergonomics, tool motion scaling, and eliminating hand tremor which significantly improves the ease of vascular anastomosis over manual techniques. In practice, however, the consistency of suture needle placement and reduction of tissue trauma remains entirely dependent on the surgeon's skill, technique, and training \cite{chen2018use}, and may be one factor why clinical outcomes and complication rates have remained relatively unchanged.  

To overcome this human-centric limitation, several autonomous robotic systems have been developed to execute surgical tasks without human intervention. For example, Knoll et al. \cite{knoll2012selective} developed a system focused on automating knot tying, and Kim et al. \cite{kim2024surgical} introduced the Surgical Robot Transformer (SRT), which can autonomously perform tasks like needle pickup, tissue lifting, and knot tying. Sen et al. \cite{goldberg2016automating} developed the Suture Needle Angular Positioner (SNAP) and a planning framework to automate suture placement. While these systems represent significant advancements in autonomous robotic suturing, they have not been demonstrated in the context of vascular surgery, as they do not address the complex tissue manipulation or deformations that microvessels are subjected to during anastomosis. 

The Smart Tissue Autonomous Robot (STAR) represents a significant development in autonomous robotic anastomosis, having successfully performed autonomous anastomosis on intestinal tissue \cite{AxelScienceRobotics}. Using a novel three-dimensional endoscope, and AI based tissue tracking, the STAR system could account for real-time soft tissue deformations to enable intestinal anastomosis during in vivo experiments. However, despite its advancements, technical limitations prevent STAR from being used to perform vascular anastomosis. For instance, STAR’s structured light vision system is effective at creating three-dimensional point clouds of larger structures like the intestine but lacks the resolution and point cloud accuracy needed when imaging small-scale vessels. In addition, STAR creates a single intraoperative suture plan that is only updated when millimeter-scale deformations are detected. This suture planner would not be appropriate for vascular suturing, where the target tissue is sometimes as small as a millimeter in diameter. In an effort to improve STAR for vascular anastomosis, we previously developed the Microvascular Anastomosis Positioning System (MAPS), a motorized tissue clamp and holder that was teleoperated to rotate vessels \cite{haworth2023development}. This initial prototype was employed in a scripted workflow where the MAPS and the STAR suture tool were moved to predetermined positions for suture placement on phantom tissue. However, the scripted workflow is inefficient and requires extensive calibration to ensure accurate system positioning before each experiment. Additionally, the MAPS system could only perform automated routines and was unable to conduct ex vivo experiments, as it lacked real-time tissue sensing, error correction, and the ability to autonomously manipulate vessels during the procedure.

For a robotic system to effectively perform autonomous vascular anastomosis, it must meet several critical capabilities: accurate tissue sensing to ensure proper suture placement, precise needle driving to avoid complications, real-time error detection and correction to maintain the integrity of the anastomosis, the ability to manipulate vessels for consistent suture spacing, minimal dependence on human intervention, and demonstrated clinical feasibility.  

In this paper, we describe and evaluate the Micro Smart Tissue Autonomous Robot (µSTAR), the first robotic system to meet these critical capabilities and to perform autonomous vascular anastomosis in ex vivo tissue. The following specific contributions of this work are what has enabled a significant advancement in autonomous robotic suturing: 
\begin{enumerate}
    \item The design of a novel suturing tool with integrated high resolution Optical Coherence Tomography (OCT) fiber-optic sensor and microcamera, for accurate, real-time tissue-tool interaction feedback and to detect and account for submillimeter tissue deformations.
    \item The development of an advanced tissue classification algorithm within the OCT framework, which enhances the accuracy of suture placement by differentiating between material types and ensuring the sutures are placed at the correct depth.
    \item The creation of a neural network to enable missed suture detection so that suturing mistakes are autonomously identified, corrected, and verified during the suturing routine, which is not possible with current autonomous systems. 
    \item An autonomous surgical controller and suturing workflow for tissue manipulation and suturing using multimodal imaging modalities for complete end to end anastomosis of vascular tissue. 
\end{enumerate}

To validate the efficacy of the µSTAR system, we conducted an ex vivo study comparing its performance with that of experienced surgeons. Notably, this marks the first instance of a robotic system autonomously performing vascular anastomosis on real tissue. The following sections detail the system's design, the methodology for evaluation, and the implications of our findings from the ex vivo evaluation.

\section{Materials and Methods}

\subsection{Robotic Suturing Tool with Integrated Optical Coherence Tomography (OCT) Fiber-Optic Sensor and Microcamera}

The µSTAR system employs the LBR Med manipulator, a seven-degree-of-freedom robotic arm developed by KUKA (KUKA AG, Augsburg, Germany) (Fig. \ref{fig:µSTARsystemoverview}A). Specifically designed for medical applications, this manipulator provides the necessary precision and flexibility to accurately position the suturing tool relative to the vasculature during vascular anastomosis. The LBR Med manipulator communicates with the host computer over a Local Area Network (LAN), enabling coordinated control of its movements during the procedure.

The suturing tool integrated into µSTAR is a modified version of the Endo360 laparoscopic suturing device (EndoEvolution, MA, USA). This tool operates by driving a curved needle along a fixed circular path to puncture tissue, as shown in Fig. \ref{fig:µSTARsystemoverview}B, which is crucial for achieving consistent suture placement in delicate vascular structures. Initially motorized and adapted for robotic use as part of the Smart Tissue Autonomous Robot (STAR) project \cite{AxelScienceRobotics}, the Endo360 has been further enhanced with a custom 3D-printed housing that integrates both an Optical Coherence Tomography (OCT) sensor and a microcamera (OV6946, Omnivision, Santa Clara, USA). This housing ensures that the OCT sensor and microcamera are rigidly fixed relative to the needle, maintaining consistent positioning during the suturing process. The motorized suturing tool is controlled by the host computer over a Controlled Area Network (CAN), facilitating precise and synchronized operation.

The OCT system provides real-time feedback on the position of the suturing tool relative to the tissue (Fig. \ref{fig:µSTARsystemoverview}B). It uses a common-path design with a single-mode fiber (1060XP, Thorlabs, Newton, NJ, USA) connected to a high-speed swept-source OEM engine (AXSUN, Billerica, MA, USA). The OCT system acquires high-resolution depth profiles at a rate of 100 kHz, capturing interference signals at the interface between the tissue and the fiber’s external medium. Communication with the host computer is established via a Transaction Processing System (TPS) connection, allowing for seamless data integration into the control system.

The microcamera is positioned to observe the suture site before and after each suture is placed (Fig. \ref{fig:µSTARsystemoverview}B). Images captured by the camera are analyzed by the system to detect missed sutures, allowing the robot to autonomously determine if a suture needs to be repeated. The images are transmitted to the host computer via a Decklink Blackmagic Frame Grabber (Blackmagic, South Melbourne, Australia), ensuring low-latency data transfer.

This novel suturing tool, with integrated OCT and microcamera, provides the essential sensing and visualization capabilities required for performing vascular anastomosis autonomously.

\subsection{Vessel Manipulation and Suturing Workflow} 

\begin{figure}[t]
    \centering
    \includegraphics[width=1\linewidth]{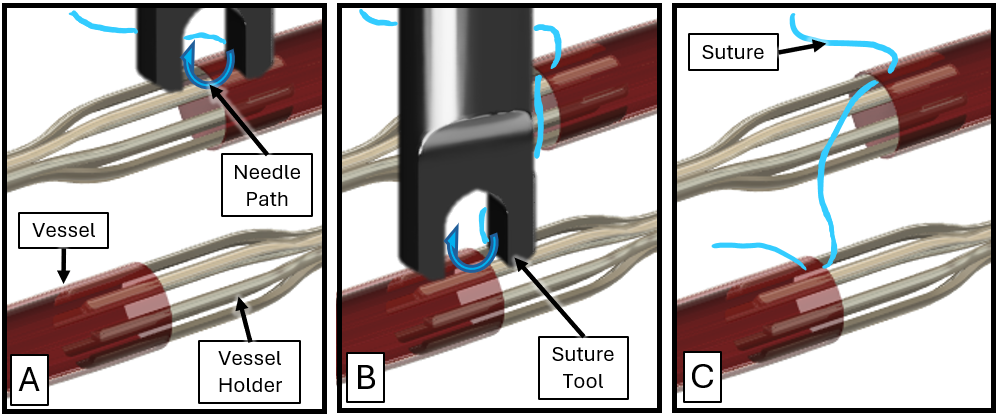}
    \caption{Example workflow for µSTAR to place a suture for anastomosis. A: Suture tool first drives the needle outside-inside in the right vessel. B: Suture tool drives the needle inside-outside in the left vessel. C: Resultant suture placed for anastomosis. The vessel can now be rotated for the next stitch.}
    \label{fig:exampleworkflow}
    \vspace{-1em}
\end{figure}

The µSTAR system incorporates the Microvascular Anastomosis Positioning System (MAPS), a manually controlled robotic manipulator designed specifically for precise vessel handling during vascular anastomosis (see Fig. \ref{fig:µSTARsystemoverview}C). Originally developed as an initial prototype for phantom tissue \cite{haworth2023development}, MAPS has been upgraded to function with real tissue by integrating a more robust control system and mechanical improvements to support the autonomous operation of the µSTAR system.

In traditional manual suturing, surgeons often use an approximator clamp to hold the vessels in place, securing them and temporarily blocking blood flow. This allows the surgeon to suture one side, flip the clamp, and suture the other side. The MAPS system emulates this clinical practice by precisely rotating the vessels during robotic suturing, ensuring proper access to all sides while minimizing the risk of excessive torque on the vessels. The system features clamp carriages that rotate in tandem with the nitinol vessel holders, ensuring synchronized movement and maintaining the alignment of the vessels throughout the procedure (see Fig. \ref{fig:µSTARsystemoverview}C).

One of the key modifications to the MAPS system is the integration of a serial interface, enabling the system to communicate with the host computer and be controlled by the µSTAR's autonomous controller. This upgrade replaces the manual control interface used in the initial prototype, making MAPS more versatile and better suited for integration into an autonomous workflow. The nitinol vessel holders, which securely grip the vessels, have been enhanced with an epoxy antislip coating (McMaster-Carr, Elmhurst, IL) to increase friction and prevent slippage during the rotation of ex vivo tissue, which is critical for maintaining alignment and accuracy during suturing (see Fig. \ref{fig:µSTARsystemoverview}B). Additionally, lubrication was added to the internal mechanisms of MAPS to ensure smoother operation, and improvements were made to the positioning control code to prevent overshooting of the target angle, further enhancing the system's precision.

To load the vessels into the MAPS system, the nitinol vessel holders are first sheathed, allowing the vessels to be easily loaded (Fig. \ref{fig:µSTARsystemoverview}C). Once the vessels are in place, the sheaths are retracted, causing the nitinol to expand inside the vessels and securely grip them. This secure grip is essential for maintaining alignment and preventing slippage during the suturing process.

In the suturing workflow, the MAPS system plays a crucial role in aligning the vessels for each suture. As illustrated in Fig. \ref{fig:exampleworkflow}, the suturing tool first drives the needle through the tissue in the right vessel half, taking the needle and suture from outside the vessel to inside the vessel (Fig. \ref{fig:exampleworkflow}A). The tool then moves to the left vessel half, driving the needle from inside to outside (Fig. \ref{fig:exampleworkflow}B). After the suture is pulled through and cut, MAPS rotates the vessel to the next suturing location (Fig. \ref{fig:exampleworkflow}C). This process ensures consistent suture placement with minimal human intervention while adhering to clinical practices.

Unlike the initial prototype, the current version of MAPS is designed to fully integrate with the autonomous control architecture of µSTAR and to handle real tissue. The modifications, including the epoxy coating on the vessel holders, improved lubrication, and enhanced control code, ensure that the system can meet the demands of vascular anastomosis with real tissue, representing an improvement over the initial design.

\subsection{Tissue Detection and Classification with OCT}

The µSTAR system utilizes an Optical Coherence Tomography (OCT) sensor to provide real-time feedback for tissue detection and classification during vascular anastomosis. This feedback is crucial for maintaining a precise bite depth, the distance from the vessel edge where the suture is placed, ensuring accurate and consistent suturing.

The OCT system provides an intensity profile over distance, which is used to differentiate between tissue, air, and nitinol. Before scanning, the OCT fiber is positioned at a predefined start location above the tissue, as shown in Fig. \ref{fig:µSTARsystemoverview}B. The system first identifies the tissue location by detecting a threshold in the OCT signal. The signal beyond this threshold point, precisely one millimeter past it, is saved as a template image representing the approximate thickness of the vessel wall.

\begin{figure}[t]
    \centering
    \includegraphics[width=1\linewidth]{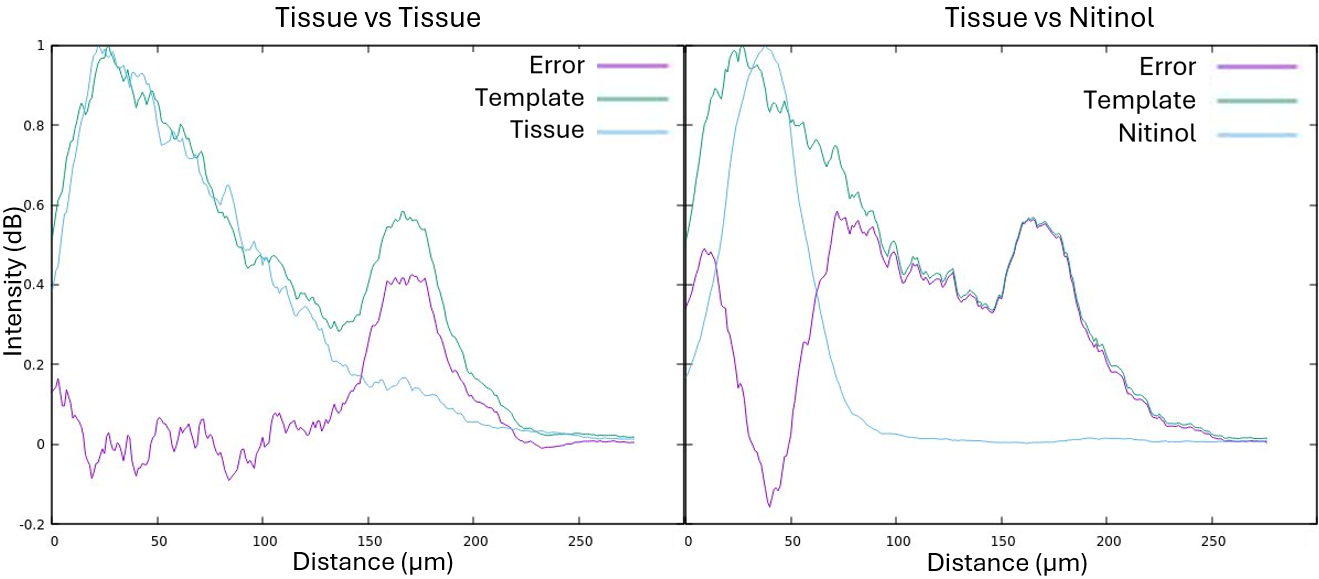}
    \caption{Example Optical Coherence Tomography (OCT) sensor signal for vessel edge detection. Tissue template compared to new tissue signal (left). Tissue template compared to nitinol signal (right).}
    \label{fig:octgraph}
\end{figure}

Before template matching is performed, the template and incoming signal are smoothed to reduce high-frequency noise in the signal. If the maximum value of the smoothed signal \( S \) is below the threshold \( \tau_{\text{air}} \), the signal is determined to be air:

\begin{equation}
\label{eq:airtresh}
\max(S) < \tau_{\text{air}} \implies \text{Signal is air}
\end{equation}

The template and the incoming signal are then normalized to ensure that the maximum intensity value is 1 for both, allowing the algorithm to be robust against signal attenuation due to factors such as distance, moisture content, or light deflection of the sample. Let \( T \) represent the template and \( S \) represent the incoming signal:

\begin{equation}
\label{eq:normalizesignal}
T_{\text{norm}} = \frac{T}{\max(T)}, \quad S_{\text{norm}} = \frac{S}{\max(S)}
\end{equation}

Next, the Root Mean Squared Error (RMSE) is calculated between the normalized template \( T_{\text{norm}} \) and every index \( i \) of the normalized signal \( S_{\text{norm}} \):

\begin{equation}
\label{eq:rmseequation}
\text{RMSE}(i) = \sqrt{\frac{1}{N} \sum_{j=1}^{N} \left(T_{\text{norm},j} - S_{\text{norm},i+j-1}\right)^2}
\end{equation}

The minimum RMSE value is then compared with a threshold \( \tau_{\text{RMSE}} \) to determine whether the signal represents tissue or nitinol:
\begin{align}
\label{eq:treshtissue}
&\min(\text{RMSE}(i)) < \tau_{\text{RMSE}} \implies \text{Signal is tissue}  \\
&\min(\text{RMSE}(i)) \geq \tau_{\text{RMSE}} \implies \text{Signal is nitinol}
\end{align}

Fig. \ref{fig:octgraph} provides an example comparison of the OCT signal, illustrating how the algorithm distinguishes between tissue and nitinol. If the algorithm identifies the material as air or nitinol, the scan is halted, and the current position of the OCT fiber is marked as the vessel edge. This process is critical for ensuring that the suture tool positions the needle at the correct bite depth relative to the vessel edge.

If the OCT edge detection fails to find an edge during the scan, it will automatically attempt another scan up to three times. If the system still fails to detect the edge after three attempts, the user must manually adjust the robot before placing the suture. Similarly, if the system incorrectly identifies a false vessel edge, the user is required to adjust the position manually before continuing the procedure.

The thresholds used in this algorithm were carefully tuned during the testing of over 24 ex vivo scans, ensuring that the system could reliably differentiate between tissue and other materials.

\subsection{Missed Suture Detection}

The missed suture detection system in the µSTAR setup plays a crucial role in identifying instances where the suture fails to engage the tissue correctly. This capability is vital for maintaining the integrity of vascular anastomosis, allowing the robotic system to autonomously correct errors in real-time, thereby enhancing the overall reliability and safety of the procedure.

We implemented missed suture detection using a ResNet-50 model \cite{resnethe2016deep}, which is well-regarded for its effectiveness in deep learning tasks while mitigating the vanishing gradient problem. This architecture was selected due to its balance between depth and computational efficiency, making it suitable for the real-time requirements of the µSTAR system.

The dataset for training the ResNet-50 model was collected using the microcamera integrated into the µSTAR system, capturing images of the suture site before and after each suture was placed, see Fig. \ref{fig:µSTARsystemoverview}B for an example image. A total of 540 image pairs were gathered, with a nearly three to one imbalance between successful and missed sutures. To address this imbalance during training, we used PyTorch’s data loader \cite{imambi2021pytorch} to ensure that each batch contained an equal number of missed and successful sutures, improving the model's learning process.

We modified the ResNet-50 model to accept a six-channel input by concatenating the RGB images from the before and after suture placements. Additionally, we added a sequential block consisting of a linear layer that reduces the output features to 230 logits, followed by batch normalization and a dropout layer with a 50\% dropout rate, before mapping to the final two output logits for binary classification.

The model was trained using PyTorch, with a cross-entropy loss function and the Adam optimizer \cite{kingma2014adam} at a learning rate of 0.0003. Data augmentation techniques, including random rotations, flips, color jitter, pixel dropout, and affine transformations, were applied to enhance the model’s robustness. Early stopping was implemented if the loss did not reduce over 30 consecutive epochs to prevent overfitting.

These training methods and model modifications aim to provide the µSTAR system with a reliable mechanism for detecting missed sutures during vascular anastomosis, enabling the robot to respond appropriately during the procedure.

\subsection{Software Architecture and Autonomous Control Workflow}

\begin{figure}[b]
    \centering
    \includegraphics[width=1\linewidth]{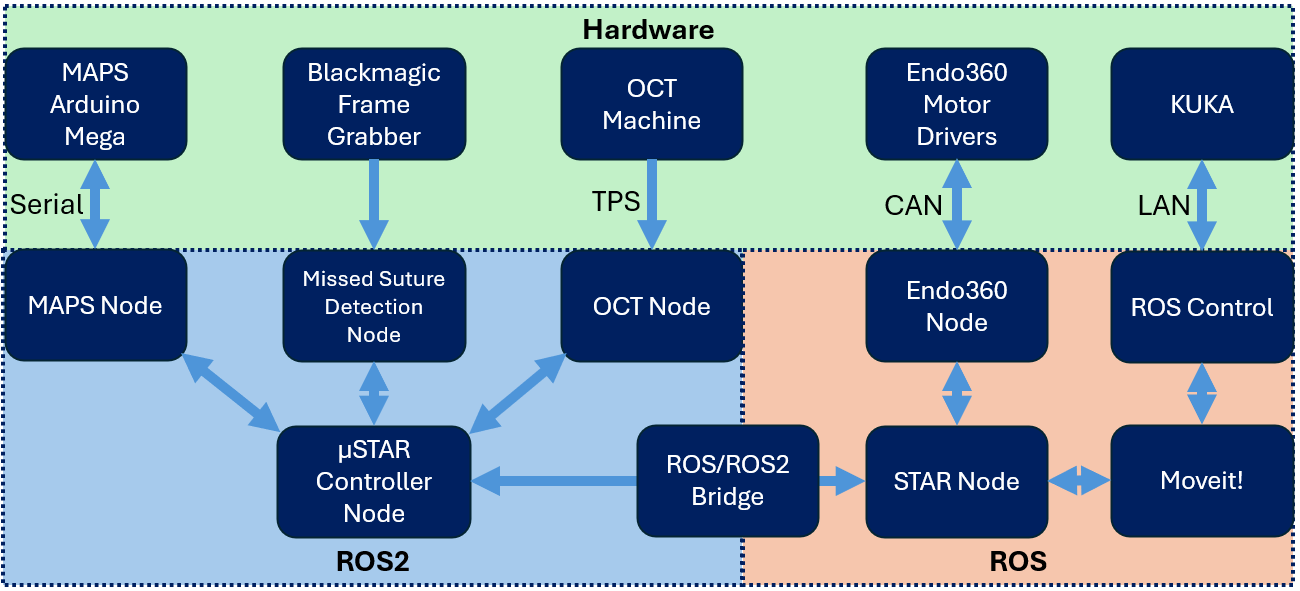}
    \caption{Software Architecture diagram for the µSTAR system. The green zone represents the hardware components, while the blue and red zones represent the ROS2 and ROS software nodes respecitively.}
    \label{fig:softwarearch}
\end{figure}

The µSTAR system integrates various hardware components using a robust software architecture to achieve autonomous vascular anastomosis. The architecture leverages both ROS (Robot Operating System) \cite{rosquigley2009ros} and ROS2 \cite{ros2macenski2022robot} frameworks to manage communication between the components, ensuring seamless operation during the procedure.

As depicted in Fig. \ref{fig:softwarearch}, the µSTAR system’s software architecture is divided into two primary frameworks: ROS and ROS2. ROS was selected for the KUKA LBR Med robotic arm and the Endo360 suturing tool because existing ROS interfaces were already available for these devices. This choice simplifies integration and leverages the mature support available in the ROS ecosystem. ROS2, on the other hand, was used for newer components such as the OCT machine, MAPS manipulator, microcamera, and the µSTAR controller. ROS2’s improved real-time performance, enhanced security, and longer-term support make it a more suitable choice for these newer components, ensuring the architecture's viability for future projects.

\begin{algorithm}[t]
\caption{µSTAR Control Workflow}
\label{alg:robot_workflow}
\begin{algorithmic}[1]
    \STATE Load Vessels \COMMENT{Performed by User}
    \STATE MAPS Rotate to Start
    \FOR{Each suture from 1 to 8}
        \STATE Capture Before Image
        \STATE Scan for Right Vessel Edge
        \STATE Place Suture in Right Vessel
        \STATE Capture After Image
        \IF{Missed Suture = True}
            \STATE Ask User If They Would Like to Try Again
        \ENDIF
        \STATE Repeat for Left Vessel
        \STATE MAPS Rotate to Next
        \STATE Pull Suture Through and Cut \COMMENT{Performed by User}
    \ENDFOR
    \STATE MAPS Rotate to Beginning
    \STATE Tie Off Suture \COMMENT{Performed by User}
\end{algorithmic}
\end{algorithm}

The µSTAR Controller Node, running on ROS2, orchestrates the overall procedure by managing and coordinating the actions of all system components. It communicates with various nodes, such as the MAPS Node, OCT Node, and Endo360 Node, through ROS services, ensuring that each node executes its respective task in the workflow. The ROS/ROS2 Bridge facilitates communication between the ROS-based components (like the KUKA arm and Endo360 tool) and the ROS2-based components. For the KUKA arm, the ROS-based STAR Node interacts with MoveIt! \cite{chitta2016moveit} for motion planning, which then interfaces with ROS Control to execute the planned movements via LAN communication. Similarly, the MAPS Node controls the MAPS manipulator via a serial interface, receiving commands from the µSTAR Controller Node to rotate the vessels during the suturing process. The OCT Node handles real-time tissue scanning, while the Endo360 Node manages the suturing tool, all within the ROS/ROS2 architecture.

The overall workflow of the µSTAR system is detailed in Algorithm \ref{alg:robot_workflow}. The process begins with the user loading the vessels into the MAPS system, followed by MAPS rotating to the start position. The µSTAR Controller then sequentially executes the suturing tasks, starting by capturing a before-suture image with the microcamera and then scanning for the right vessel edge using the OCT system. Next, the suture is placed in the right vessel using the Endo360 tool after which an after-image of the suture site is captured. If a missed suture is detected, the system prompts the user to decide whether to retry the suture. This procedure is repeated for the left vessel, after which MAPS rotates the vessel to the next suturing location.

After the vessel edge is identified by the OCT system, the robot moves a pre-set distance to align the needle relative to the vessel edge and places the suture. This pre-set distance was determined through a calibration process where the suturing tool was first moved flush against a piece of paper, and the distance measured by the OCT system was saved as the z offset. Then, the needle was used to puncture the paper, and the robot was jogged until the puncture was observed under the OCT system. The total jog distance was recorded as the x and y offsets. This calibration ensures that the needle is correctly aligned with the vessel edge during the suturing process.

By combining the strengths of ROS and ROS2, the µSTAR system ensures that all components work in harmony to perform the vascular anastomosis with precision, while maintaining flexibility for future enhancements. The use of this architecture enables the system to autonomously manage complex tasks, reducing the need for human intervention while maintaining high surgical precision.

\section{Experiments and Results}

\subsection{Vessel Positioning System Evaluation}

\subsubsection{Repeatability Testing}
The repeatability of the MAPS system was assessed by rotating a five millimeter diameter ex vivo porcine femoral artery (Animal Technologies, Tyler, TX) in 45-degree increments across the system's full range of motion. Each vessel was prepared by finding a point along the flat vessel that measured six millimeters wide with a ruler (approximately 4.5 mm diameter). It was then cut with a scalpel and a suture knot was placed on the vessel as a visual marker of its motion. Axial images were taken of the vessel before and after each rotation, and ImageJ \cite{imageJ} was used to measure the absolute angle relative to a static reference marker. The angle of each rotation was then calculated from the angular difference between each before and after image pair. 

The results of the vessel repeatability testing showed that the left stage exhibited an average rotation of 44.9 degrees (n=16) with a standard deviation of 2.8 degrees, while the right stage showed an average rotation of 45.3 degrees (n=16) with a standard deviation of 2.2 degrees. For a full rotation, the average error for both holders together was 1.9$\pm$1.3 degrees (n=4).

\begin{table}[t]
\caption{MAPS Vessel Holder Grip Force}
\label{mapsgrip}
\begin{center}
\begin{tabular}{|c|c|c|c|c|}
\hline
 & \makecell{MAX \\Axial\\Force Left \\(N)} & \makecell{MAX \\Tangent\\Force Left \\(N)} & \makecell{MAX \\Axial\\Force Right \\(N)} & \makecell{MAX \\Tangent\\Force Right \\(N)}\\
\hline
Avg. & \makecell{$0.24$\\$\pm 0.04$\\(n=3)} & \makecell{$0.25$\\$\pm 0.04$\\(n=3)} & \makecell{$0.22$\\$\pm 0.03$\\(n=3)} & \makecell{$0.25$\\$\pm 0.02$\\(n=3)} \\
\hline
Total Avg. & \multicolumn{4}{c|}{\makecell{$0.24$$\pm 0.03$ \\(n=12)}} \\
\hline
\end{tabular}
\end{center}
\vspace{-1em}
\end{table}

\begin{figure}[b]
    \vspace{-1em}
    \centering
    \includegraphics[width=0.75\linewidth]{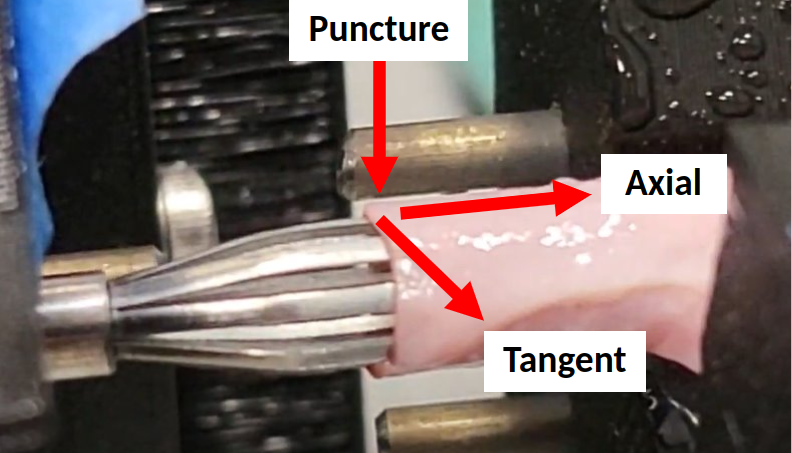}
    \caption{Diagram showing the force directions for grip strength testing. Puncture force is measured by puncturing a needle through the vessel. Axial and Tangent forces are measured by pulling suture tied at the vessel edge.}
    \label{fig:gripforcediagram}
\end{figure}

\subsubsection{Grip Force and Puncture Testing}
The grip strength of the nitinol vessel holder was evaluated by tying one end of the suture onto a five millimeter diameter porcine artery (Animal Technologies, Tyler, TX). Each vessel was prepared by finding a point along the flat vessel that measured six millimeters wide with a ruler (approximately 4.5 mm diameter), and cutting it with a scalpel. Then the vessel was loaded onto the nitinol vessel holder and the other end of the suture was attached to a force sensor (GS0-1K, Transducer Techniques, Temecula, CA) to measure the maximum forces required to cause the vessel to slip along both axial and tangential directions, as shown in Fig. \ref{fig:gripforcediagram}. The force was recorded in the axial and tangential directions three times each, which was repeated for both vessel holders. Additionally, the maximum puncture force was determined by using a 3-0 needle attached to the force sensor to puncture the vessel on the holder perpendicularly 16 times, and the maximum force was recorded.

The results of the grip force testing are summarized in Table \ref{mapsgrip}. The maximum force required to puncture the vessel was measured at 0.80 N (n=16). The total average grip force for the MAPS vessel holders was 0.24$\pm$0.03 N (n=12), indicating that a steep puncture angle could potentially cause the vessel to slip on the holder.

\subsection{Tissue Classification Testing and Evaluation}
The tissue classification capability of the µSTAR system was evaluated using five millimeter diameter ex vivo porcine arteries (Animal Technologies, Tyler, TX). Each vessel was prepared by finding a point along the flat vessel that measured six millimeters wide with a ruler (approximately 4.5 mm diameter), and cutting it with a scalpel. The arteries were loaded onto the nitinol vessel holders, and OCT scans were performed to identify the tissue edge and classify the transition material as either nitinol or air. 

A total of 49 scans were conducted across six different artery samples. The system successfully identified the tissue edge in 89.8\% of the scans. After identifying the edge, the system correctly labeled the transition material as either nitinol or air in 88.6\% of the samples.

\subsection{Missed Suture Detection Evaluation}

\begin{figure}[b]
    \centering
    \includegraphics[width=1\linewidth]{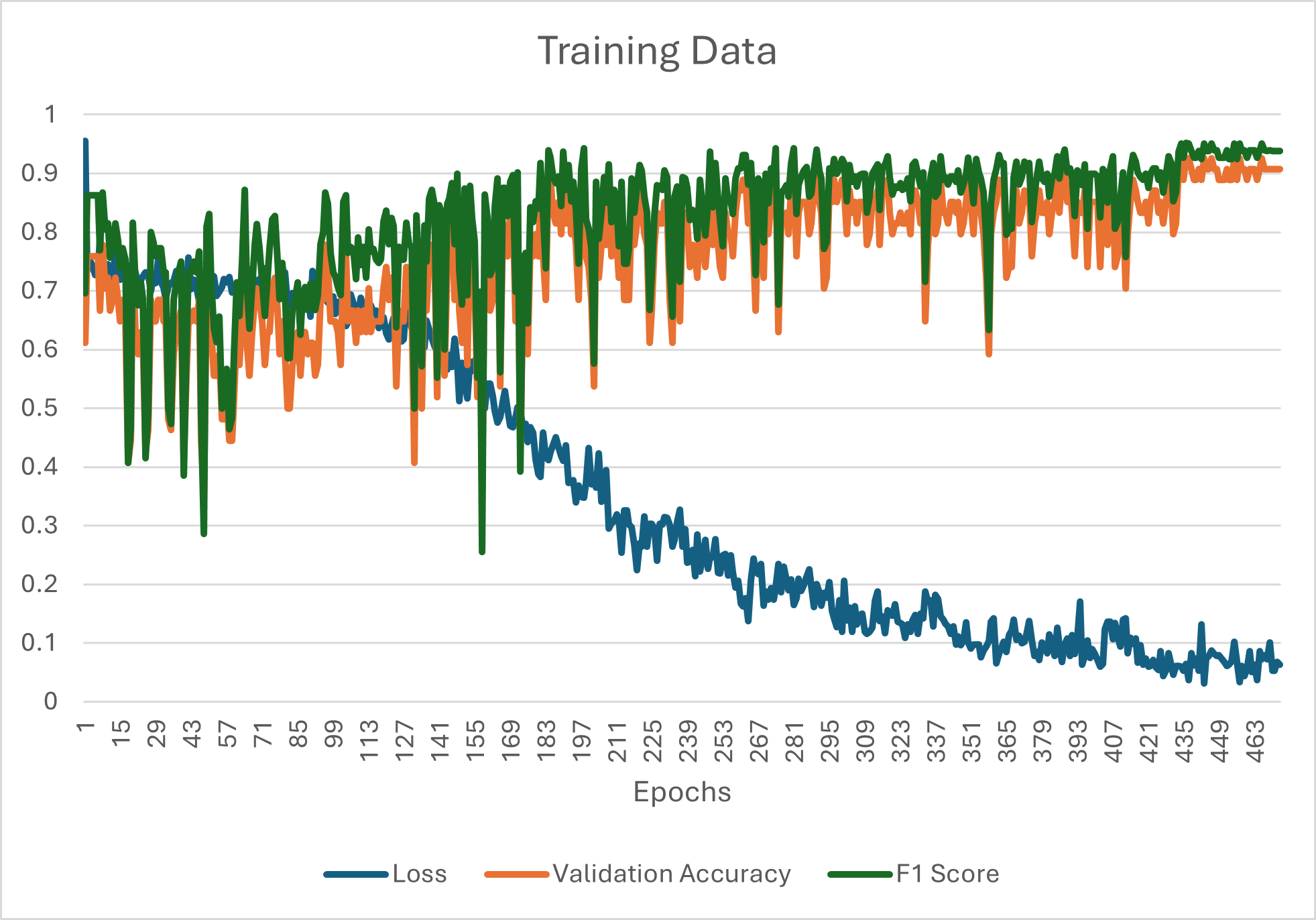}
    \caption{Training data for the missed suture detection ResNet50 model showing Loss, Validation Accuracy, and F1 Score over training epochs.}
    \label{fig:cvtraininggraph}
\end{figure}

The effectiveness of our custom ResNet50 model for missed suture detection was assessed by analyzing its performance across the training, validation, and test datasets. The model was trained using a dataset of 540 image pairs, which were divided into 432 training pairs, 54 validation pairs, and 54 test pairs.

The training progress of the model is illustrated in Fig. \ref{fig:cvtraininggraph}, which shows the Loss, Validation Accuracy, and F1 Score over the training epochs. The model's loss consistently decreased throughout the training process, indicating effective learning and optimization. The validation accuracy and F1 score exhibited fluctuations in the early stages of training, as the model adjusted its parameters. However, as training progressed, these metrics stabilized, reflecting the model's ability to generalize well on unseen data.

Upon completion of training, the model achieved an accuracy of 96.5\% on the training data, 90.74\% on the validation data, and 87.04\% on the test data. These results suggest that while the model performed well on the training and validation datasets, there is room for improvement in its generalization to the test set. The model’s F1 Score, which is particularly useful for evaluating binary classification tasks with imbalanced datasets, remained high, further demonstrating the model's capability to accurately distinguish between successful and missed sutures.

\subsection{Ex Vivo Comparison Study}

To evaluate the performance of the µSTAR system in a clinically relevant setting, an ex vivo comparison study was conducted using five millimeter diameter porcine artery (Animal Technologies, Tyler, TX). Three surgeons, each with different specialized training in head and neck surgery, oral and maxillofacial surgery, and trauma surgery, respectively, participated in the study. Each surgeon and the µSTAR system performed anastomosis on five porcine femoral arteries, with one additional sample used for practice.

\begin{figure}[b]
    \centering
    \includegraphics[width=0.75\linewidth]{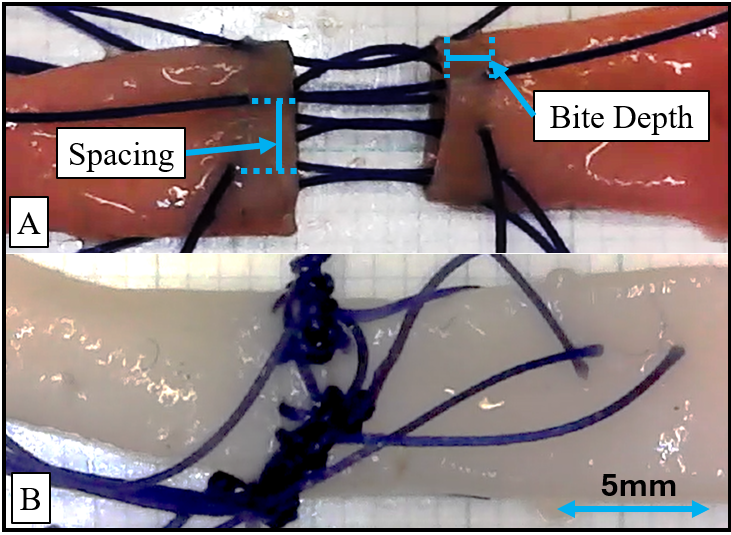}
    \caption{A: Anastomosis performed by µSTAR before knots are tied using 3-0 suture. B: Anastomosis after knots are tied by non-surgeon.}
    \label{fig:anastomosiscloseup}
\end{figure}

During the experiments, each surgeon selected a portion of the porcine femoral artery approximately five millimeters in diameter using a ruler, cut the vessel with a scalpel, and performed the anastomosis using interrupted sutures. The µSTAR system followed a similar procedure, with a graduate student tying off the sutures after the robot had finished placing the sutures. An example anastomosis performed by µSTAR can be seen in Fig. \ref{fig:anastomosiscloseup}. The surgeons used 6-0 polypropylene monofilament suture, while the µSTAR system utilized 3-0 braided polyester suture, the smallest size compatible with the µSTAR suturing tool. The average time per stitch was recorded by timing from the needle's first contact with the tissue to the final knot being tied and cut, then dividing the total procedure time by the number of sutures placed. 

Immediately after suturing was complete, each sample was tested for bubble leak. Bubble leak testing was performed by attaching a pressure gauge (0.05\% Test Gauge, Ashcroft, Stratford, CT) to an indeflator and a needle-dispensing tip. The dispensing tip was clamped into one end of the vessel, while the opposite end was also clamped off. The sample was then submerged in water and pressurized until bubbles were observed coming from the anastomosis, at which point the corresponding pressure was recorded.

Following the bubble leak test, the lumen of the anastomosis was measured using 3D printed pin gauges in 0.5 mm diameter increments. The inner diameter (ID) was measured at the anastomosis site and compared to the lumen of the vessel just outside the anastomosis. The percentage reduction in lumen diameter was calculated using Equation \ref{eq:lumenred}, reflecting the extent to which the anastomosis constricts blood flow. 

\begin{equation}
\label{eq:lumenred}
\% Reduction = 100*(1 - (\frac{Anastomosis ID}{Raw Vessel ID})^2)
\end{equation}

After the lumen measurement, the samples were untied, and measurements were taken of the suture bite depth and suture spacing using ImageJ \cite{imageJ}. Bite depth refers to the distance of the suture placement from the edge of the vessel, while suture spacing refers to the distance between consecutive sutures as shown in Fig. \ref{fig:anastomosiscloseup}A. For the µSTAR system, a bite depth of 1.5 mm was chosen, as it is 1.5 times the vessel wall thickness of one millimeter \cite{zheng2020large}. To assess the consistency of suture placement, the coefficient of variance (COV\%) was calculated for both bite depth and suture spacing using Equation \ref{eq:cov}:

\begin{equation}
\label{eq:cov}
COV\% = (\frac{Standard Deviation}{Average}) * 100
\end{equation}

COV\% was selected as a metric because it is agnostic to the surgeon’s chosen number of sutures or target bite depth, allowing for an unbiased comparison of how much the suture placement varied within each group.

\begin{table}[t]
\caption{µSTAR and Surgeon Ex Vivo Average Results}
\label{ustarandsurgeonresults}
\begin{center}
\begin{tabular}{|c|c|c|c|c|c|}
\hline
 Surgeon & \makecell{Bite\\Depth\\COV\%} & \makecell{Suture\\ Spacing\\COV\%} & \makecell{Lumen\\Reduction\\(\%)}  & \makecell{Bubble\\Leak\\(PSI)} & \makecell{Avg. Time\\Per Stitch\\(seconds)}\\
\hline
1 & \makecell{$35$\\(n=75)}& \makecell{$27$\\(n=74)}&  \makecell{$21$ \\ $\pm 11$\\(n=6)} &  \makecell{$0.34$ \\ $\pm 0.13$\\(n=6)} &  \makecell{$90$ \\ $\pm 22$\\(n=5)}\\
\hline
2 & \makecell{$42$\\(n=58)}& \makecell{$62$\\(n=58)}&  \makecell{$71$ \\ $\pm 28$\\(n=5)} &  \makecell{$0.38$ \\ $\pm 0.06$\\(n=4)} &  \makecell{$158$ \\ $\pm 55$\\(n=5)}\\
\hline
3 & \makecell{$34$\\(n=80)}& \makecell{$35$\\(n=80)}&  \makecell{$39$ \\ $\pm 29$\\(n=5)} &  \makecell{$0.32$ \\ $\pm 0.11$\\(n=5)} &  \makecell{$176$ \\ $\pm 27$\\(n=5)}\\
\hline
µSTAR  & \makecell{$33$\\(n=80)}& \makecell{$30$\\(n=80)}&  \makecell{$26$\\$\pm 17$\\(n=5)} &  \makecell{$0.32$\\$\pm 0.23$\\(n=5)} &  \makecell{$353$\\$\pm 40$\\(n=5)}\\
\hline
\end{tabular}
\end{center}
\end{table}

During the ex vivo study, the LBR Med manipulator lost connection five times, requiring reconnection. Once reconnected, the procedure was able to continue as normal, but this issue will need to be addressed in future revisions. The time when the manipulator was disconnected was subtracted from µSTAR's total procedure time. 

The µSTAR system successfully completed the anastomoses, with 90\% of the sutures placed without human intervention. The resulting bite depth was an average of 1.54$\pm$0.22 mm, with an average error of 0.39 mm from the target 1.5 mm bite depth. Some slipping of the vessel along the axis of the nitinol vessel holder was observed during the stitch placement, likely caused by the curved path of the needle through the tissue. Additionally, the vessel was observed to slip about the holder axis later in the procedure when the tangential tension of the suspended suture increased. This led to one instance of a crossed stitch as the vessel rotated, causing the placement of the final suture to overlap with the first suture.

\begin{figure}[t]
    \centering
    \includegraphics[width=1.00\linewidth]{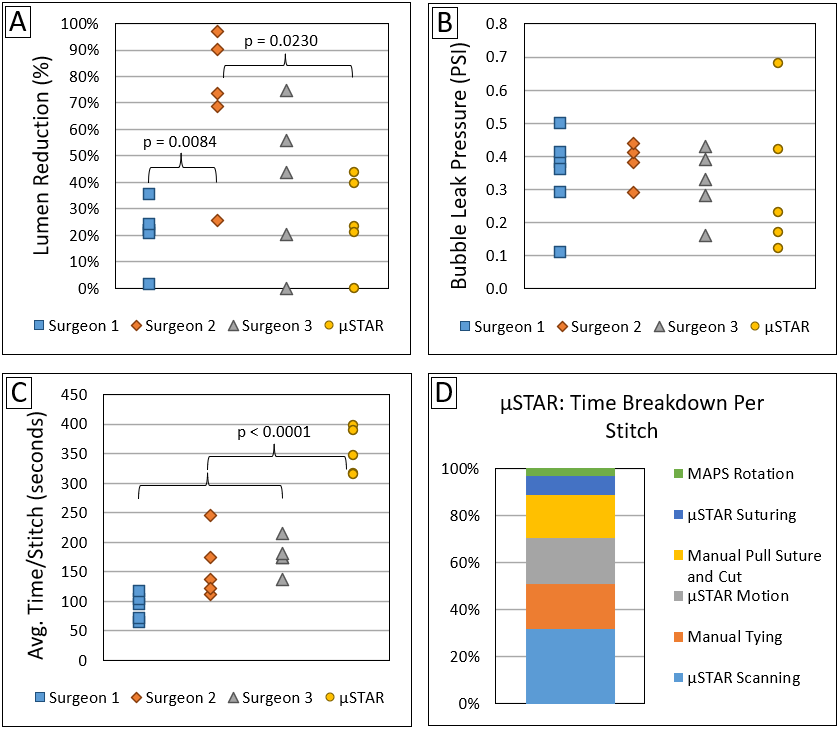} % Replace with your image file
    \caption{Graphs comparing the ex vivo results of µSTAR and the surgeons for A: Lumen Reduction, B: Bubble Leak, and C: Average Time Per Stitch. Panel D shows the time per stitch breakdown for the µSTAR system. Only significant (<0.05) p-values are listed.}
    \label{fig:allgraphs}
\end{figure}

Among the measured outcomes, there were no statistically significant differences between the individual surgeons and the µSTAR system for bubble leak (Fig. \ref{fig:allgraphs}B). However, for lumen reduction and time per stitch, statistically significant differences were observed within the group (Figs. \ref{fig:allgraphs}A and C). Notably, µSTAR outperformed Surgeon 2 in terms of suture spacing COV\%, with µSTAR achieving a COV\% of 30\% compared to Surgeon 2's 62\% (Table \ref{ustarandsurgeonresults} and Fig. \ref{fig:COVgraphs}B).

The µSTAR system was significantly slower than all three surgeons, with the average time per stitch being 352.7 seconds, compared to an average of 141.2 seconds for the surgeons. A breakdown of the time per stitch for µSTAR is shown in Fig. \ref{fig:allgraphs}D, indicating that the majority of the procedure time was due to scanning for the vessel edge and moving between points.

\subsection{Statistical Analysis}
\begin{figure}[t]
    \centering
    \includegraphics[width=1.00\linewidth]{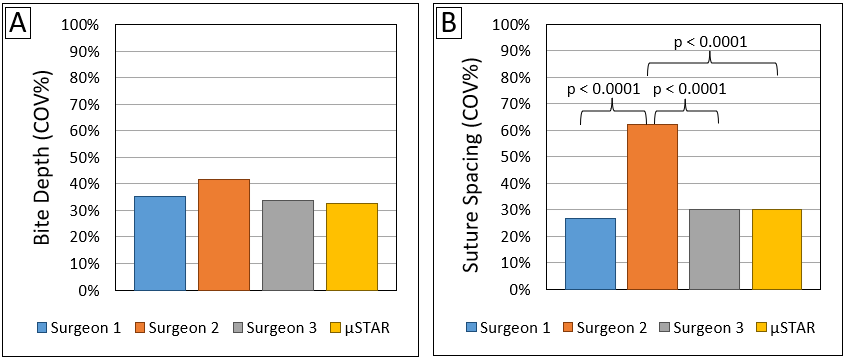} % Replace with your image file
    \caption{Graphs comparing the ex vivo results for the µSTAR system and the surgeons for A: Bite Depth COV\% and B: Suture Spacing COV\%. Only significant (<0.05) p-values are listed. }
    \label{fig:COVgraphs}
\end{figure}

Statistical analysis was performed in Microsoft Excel using the Real Statistics Resource Pack (Release 9.1.1; 2024, Charles Zaiontz). A one-way analysis of variance (ANOVA) was used to compare the surgeon's individual performances and that of the µSTAR system. This method allows for comparison of multiple groups while reducing Type I errors. Any group p-value below the level of significance (0.05) was then subjected to a Post-Hoc analysis using Tukey HSD. Statistically significant results are labeled in the graphs in Figs. \ref{fig:COVgraphs} A, B, and C. Statistical differences for COV\% were calculated using the MedCalc online calculator \cite{forkman2009estimator} and are reported in Figs. \ref{fig:COVgraphs}A and B.

\section{Discussion}

The µSTAR system represents an important development in the field of autonomous robotic vascular surgery, particularly in the challenging domain of anastomosis. The ex vivo comparison study demonstrated that µSTAR can perform anastomosis with precision and consistency, achieving outcomes that are competitive with those of experienced surgeons. This is a noteworthy achievement, considering the complexity of the task and the reliance on surgeon skill in traditional methods.

One of the key strengths of the µSTAR system is its ability to perform the majority of the suturing process autonomously, with 90\% of the sutures placed without human intervention. This level of autonomy is unprecedented in vascular surgery and highlights the potential of autonomous surgical systems to reduce dependence on individual surgeon skill. The system’s integration of OCT for real-time tissue sensing and a ResNet50-based missed suture detection algorithm further enhances its capability, allowing for precise suture placement and error correction during the procedure.

However, several limitations of the µSTAR system were identified during the study. One significant limitation is the use of a 3-0 braided polyester suture, which is larger than the sutures typically used for this size of anastomosis. This was necessitated by the design of the µSTAR suturing tool, which currently does not support smaller suture sizes. Despite this limitation, the µSTAR system remained competitive in terms of lumen reduction, demonstrating its effectiveness even with larger sutures. Future iterations of the µSTAR system should aim to support smaller suture sizes to align more closely with clinical practices in microvascular surgery.

Another limitation is the manual knot-tying performed by a graduate student after the robot completed the suturing process. This step introduces variability in the final anastomosis, particularly in the bubble leak test, where inconsistencies in knot-tying could affect the leak pressure results. Future iterations of the µSTAR system should aim to automate the knot-tying process to reduce this variability and further enhance the system’s autonomy.

Additionally, while the system was able to perform suturing with a high degree of precision, some issues with tissue manipulation were observed. Slippage of the vessel along the axis of the nitinol vessel holder occurred during the procedure, likely due to the curved path of the needle through the tissue. This slippage, along with the rotational movement of the vessel holder, led to a crossed stitch in one instance. Addressing these issues will be crucial for improving the consistency of suture placement in future versions of the system.

The µSTAR system was notably slower than the surgeons, primarily due to the time required for robot movement, especially during the scanning for the vessel edge. In future iterations, speeding up the robot’s movements could bring the procedure time closer to that of human surgeons, making the system more efficient without sacrificing precision. Additionally, the manual knot-tying by a graduate student contributed to the overall time, suggesting that automating this process could further reduce the procedure duration. 

Currently, due to the size of the suturing tool and the nitinol vessel holders, the µSTAR system is limited to performing anastomosis on vessels approximately five millimeters in diameter. Future work will focus on miniaturizing the system to enable the anastomosis of smaller vessels.

\section{Conclusion}

The development and successful performance of the µSTAR system in this study mark an important advancement toward fully autonomous vascular anastomosis. µSTAR has shown the ability to perform vascular anastomosis autonomously with results competitive to that of experienced surgeons and successfully placing 90\% of sutures without human intervention. This represents a significant step in robotic vascular surgery, potentially reducing reliance on highly skilled surgeons for complex procedures. However, the study also identified areas for improvement, such as the need for compatibility with smaller sutures and fully autonomous knot-tying. Future efforts will focus on these enhancements, aiming to further refine µSTAR’s capabilities and miniaturize the system for microvascular surgery on smaller, more challenging vessels.

\bibliographystyle{IEEEtran}
\bibliography{ref}

\begin{thebibliography}{10}
\providecommand{\url}[1]{#1}
\csname url@samestyle\endcsname
\providecommand{\newblock}{\relax}
\providecommand{\bibinfo}[2]{#2}
\providecommand{\BIBentrySTDinterwordspacing}{\spaceskip=0pt\relax}
\providecommand{\BIBentryALTinterwordstretchfactor}{4}
\providecommand{\BIBentryALTinterwordspacing}{\spaceskip=\fontdimen2\font plus
\BIBentryALTinterwordstretchfactor\fontdimen3\font minus \fontdimen4\font\relax}
\providecommand{\BIBforeignlanguage}[2]{{%
\expandafter\ifx\csname l@#1\endcsname\relax
\typeout{** WARNING: IEEEtran.bst: No hyphenation pattern has been}%
\typeout{** loaded for the language `#1'. Using the pattern for}%
\typeout{** the default language instead.}%
\else
\language=\csname l@#1\endcsname
\fi
#2}}
\providecommand{\BIBdecl}{\relax}
\BIBdecl

\bibitem{PlasticSurgeryStats2020}
``National plastic surgery statistics,'' American Society of Plastic Surgeons, Tech. Rep., 2020.

\bibitem{heidekrueger2022comparison}
P.~Heidekrueger \emph{et~al.}, ``Comparison of venous couplers versus hand-sewn technique in 4577 cases of diep-flap breast reconstructions--a multicenter study,'' \emph{Microsurgery}, vol.~42, no.~1, pp. 5--12, 2022.

\bibitem{goss2018fundamentals}
S.~G. Goss and D.~M. Salvatore, ``Fundamentals of vascular anastomosis,'' \emph{Fundamentals of General Surgery}, pp. 239--252, 2018.

\bibitem{saade2023tricks}
F.~Saade, A.~Quemener-Tanguy, L.~Obert, S.~El-Rifai, C.~Bouteille, and F.~Loisel, ``Tricks in end-to-end anastomosis in microsurgery: a systematic review,'' \emph{Indian Journal of Surgery}, vol.~85, no.~4, pp. 740--747, 2023.

\bibitem{aamcglobaldata2024}
G.~Plc, ``The complexities of physician supply and demand: Projections from 2021 to 2036,'' AAMC, Washington, DC, Tech. Rep., 2024.

\bibitem{lai2019robot}
C.-S. Lai, C.-T. Lu, S.-A. Liu, Y.-C. Tsai, Y.-W. Chen, and I.-C. Chen, ``Robot-assisted microvascular anastomosis in head and neck free flap reconstruction: preliminary experiences and results,'' \emph{Microsurgery}, vol.~39, no.~8, pp. 715--720, 2019.

\bibitem{SymaniInVivoTrial2023}
G.~Malzone, G.~Menichini, M.~Innocenti, and A.~Ballest{\'\i}n, ``Microsurgical robotic system enables the performance of microvascular anastomoses: a randomized in vivo preclinical trial,'' \emph{Scientific Reports}, vol.~13, no.~1, p. 14003, 2023.

\bibitem{micosuremusatrial2022}
T.~J. van Mulken \emph{et~al.}, ``One-year outcomes of the first human trial on robot-assisted lymphaticovenous anastomosis for breast cancer--related lymphedema,'' \emph{Plastic and Reconstructive Surgery}, vol. 149, no.~1, pp. 151--161, 2022.

\bibitem{chen2018use}
J.~Chen, P.~J. Oh, N.~Cheng, A.~Shah, J.~Montez, A.~Jarc, L.~Guo, I.~S. Gill, and A.~J. Hung, ``Use of automated performance metrics to measure surgeon performance during robotic vesicourethral anastomosis and methodical development of a training tutorial,'' \emph{The Journal of urology}, vol. 200, no.~4, pp. 895--902, 2018.

\bibitem{knoll2012selective}
A.~Knoll, H.~Mayer, C.~Staub, and R.~Bauernschmitt, ``Selective automation and skill transfer in medical robotics: a demonstration on surgical knot-tying,'' \emph{The International Journal of Medical Robotics and Computer Assisted Surgery}, vol.~8, no.~4, pp. 384--397, 2012.

\bibitem{kim2024surgical}
J.~W. Kim \emph{et~al.}, ``Surgical robot transformer (srt): Imitation learning for surgical tasks,'' \emph{arXiv preprint arXiv:2407.12998}, 2024.

\bibitem{goldberg2016automating}
S.~Sen, A.~Garg, D.~V. Gealy, S.~McKinley, Y.~Jen, and K.~Goldberg, ``Automating multi-throw multilateral surgical suturing with a mechanical needle guide and sequential convex optimization,'' in \emph{2016 IEEE international conference on robotics and automation (ICRA)}.\hskip 1em plus 0.5em minus 0.4em\relax IEEE, 2016, pp. 4178--4185.

\bibitem{AxelScienceRobotics}
H.~Saeidi \emph{et~al.}, ``Autonomous robotic laparoscopic surgery for intestinal anastomosis,'' \emph{Science robotics}, vol.~7, no.~62, p. eabj2908, 2022.

\bibitem{haworth2023development}
J.~Haworth \emph{et~al.}, ``Development and evaluation of a robotic vessel positioning system for semi-automatic microvascular anastomosis,'' in \emph{2023 IEEE International Conference on Robotics and Automation (ICRA)}.\hskip 1em plus 0.5em minus 0.4em\relax IEEE, 2023, pp. 6901--6908.

\bibitem{resnethe2016deep}
K.~He, X.~Zhang, S.~Ren, and J.~Sun, ``Deep residual learning for image recognition,'' in \emph{Proceedings of the IEEE conference on computer vision and pattern recognition}, 2016, pp. 770--778.

\bibitem{imambi2021pytorch}
S.~Imambi, K.~B. Prakash, and G.~Kanagachidambaresan, ``Pytorch,'' \emph{Programming with TensorFlow: solution for edge computing applications}, pp. 87--104, 2021.

\bibitem{kingma2014adam}
D.~P. Kingma and J.~Ba, ``Adam: A method for stochastic optimization,'' \emph{arXiv preprint arXiv:1412.6980}, 2014.

\bibitem{rosquigley2009ros}
M.~Quigley \emph{et~al.}, ``Ros: an open-source robot operating system,'' in \emph{ICRA workshop on open source software}, vol.~3.\hskip 1em plus 0.5em minus 0.4em\relax Kobe, Japan, 2009, p.~5.

\bibitem{ros2macenski2022robot}
S.~Macenski, T.~Foote, B.~Gerkey, C.~Lalancette, and W.~Woodall, ``Robot operating system 2: Design, architecture, and uses in the wild,'' \emph{Science robotics}, vol.~7, no.~66, p. eabm6074, 2022.

\bibitem{chitta2016moveit}
S.~Chitta, ``Moveit!: an introduction,'' \emph{Robot Operating System (ROS) The Complete Reference (Volume 1)}, pp. 3--27, 2016.

\bibitem{imageJ}
C.~A. Schneider, W.~S. Rasband, and K.~W. Eliceiri, ``Nih image to imagej: 25 years of image analysis,'' \emph{Nature methods}, vol.~9, no.~7, pp. 671--675, 2012.

\bibitem{zheng2020large}
Y.~D. Zheng \emph{et~al.}, ``Large and uneven bites in end-to-end anastomosis of the rat femoral artery,'' \emph{Journal of Reconstructive Microsurgery}, vol.~36, no.~07, pp. 486--493, 2020.

\bibitem{forkman2009estimator}
J.~Forkman, ``Estimator and tests for common coefficients of variation in normal distributions,'' \emph{Communications in Statistics—Theory and Methods}, vol.~38, no.~2, pp. 233--251, 2009.

\end{thebibliography}

\vfill

\end{document}